%% file: main.tex
\documentclass{article}

\PassOptionsToPackage{numbers, compress}{natbib}

     \usepackage[preprint]{neurips_2019}

\usepackage[utf8]{inputenc} 
\usepackage[T1]{fontenc}    
\usepackage{hyperref}       
\usepackage{url}            
\usepackage{booktabs}       
\usepackage{amsfonts}       
\usepackage{nicefrac}       
\usepackage{microtype}      
\usepackage{graphicx}
\usepackage{caption}
\usepackage{subcaption}
\usepackage{verbatim}
\usepackage{multirow}

\newcounter{prob}
\newtheorem{problem}[prob]{Problem}

\newcommand{\squishlist}{
   \begin{list}{$\bullet$}
    { \setlength{\itemsep}{0pt}
      \setlength{\parsep}{2pt}
      \setlength{\topsep}{2pt}
      \setlength{\partopsep}{0pt}
    }
}

\newcommand{\squishend}{\end{list}}

\title{Neural Input Search for Large Scale Recommendation Models}

\author{%
  Manas R.~Joglekar\\
  Google\\
  Mountain View, CA 94043\\
  \texttt{manasrj@google.com} \\
  \And
  Cong Li\\
  Google\\
  Mountain View, CA 94043\\
  \texttt{congcli@google.com} \\
  \And
  Jay K.~Adams\\
  Google\\
  Mountain View, CA 94043\\
  \texttt{jka@google.com} \\
  \And
  Pranav Khaitan\\
  Google\\
  Mountain View, CA 94043\\
  \texttt{pranavkhaitan@google.com} \\
  \And
  Quoc V.~Le\\
  Google\\
  Mountain View, CA 94043\\
  \texttt{qvl@google.com}
}

\begin{document}

\maketitle

\begin{abstract}
\input{abstract}
\end{abstract}

\section{Introduction}
\label{intro}
\input{introduction}

\section{Related Work}
\input{related_work}

\section{Neural Input Search}
\input{neural_input_search}

\section{Experiments}
\input{experiments}

\section{Conclusion}
\input{conclusion}

\bibliographystyle{abbrv}
\bibliography{reference}

\end{document}

%% file: abstract.tex

Recommendation problems with large numbers of discrete items, such as products,
webpages, or videos, are ubiquitous in the technology industry. Deep neural
networks are being increasingly used for these recommendation problems. These
models use \textit{embeddings} to represent discrete items as
continuous vectors, and the vocabulary sizes and embedding dimensions, although heavily influence the
model's accuracy, are often manually selected in a heuristical manner. We present Neural Input Search
(NIS), a technique for learning the optimal vocabulary sizes and embedding dimensions for
categorical features. The goal is to maximize prediction accuracy subject to a
constraint on the total memory used by all embeddings.
Moreover, we argue that the traditional Single-size Embedding (SE), which uses the same embedding dimension
for all values of a feature, suffers from inefficient usage of model capacity and training data.
 We propose a novel type of embedding, namely Multi-size Embedding
(ME), which allows the embedding dimension to vary for different values of the
feature. During training we use reinforcement learning to find the optimal
vocabulary size for each feature and embedding dimension for each
value of the feature. In experiments on two common types of large scale recommendation problems, i.e.
retrieval and ranking problems, NIS automatically found better vocabulary and embedding sizes
that result in $6.8\%$ and $1.8\%$ relative improvements on Recall@1 and ROC-AUC over manually optimized ones.

%% file: introduction.tex

Most modern neural network models can be thought of as comprising two
components: an input component that converts raw (possibly categorical) input data into floating point values;
and a representation learning component that combines the outputs of the
input component and computes the final output of the model.
Designing neural network architectures in an automated, data driven manner
(\textit{AutoML}) has recently attracted a lot of research interest, since the
publication of \cite{zoph2017}. However, previous research in this area has primarily focused on automated
design of the representation learning component, and little attention has been paid to the
input component. This is because most research
has been conducted on image understanding problems~\cite{enas,zoph2018,snas,liu2018darts}, where the
representation learning component is very important to model
performance, while the input component is trivial since the image pixels are already in floating point form.

For large scale recommendation problems commonly encountered in industry, the
situation is quite different. While the representation learning component is
important, the input component plays an even more critical role in the
model. This is because many recommendation problems
involve categorical features with large cardinality, and the input component assigns embedding vectors
to each item of these discrete features. This results in a huge number of embedding parameters in the input component,
which dominate both the size and the inductive bias of the model. For example, the YouTube video recommendation model
(\cite{youtube_paper})
uses a video ID vocabulary of size 1 million, with 256 dimensional embedding
vectors for each ID. This means 256 million parameters are used
just for the video ID feature, and the number grows quickly as more discrete
features are added. In contrast,
the representation learning component consists of only
three fully connected layers. So the number of model parameters is heavily
concentrated in the input component, which naturally has high impact on
model performance. In practice, despite their importance, vocabulary and embedding sizes for discrete features are often selected heuristically,
by trying out many models with different manually crafted configurations.
Since these models are usually large and expensive to train, such an approach is computationally intensive and may result in suboptimal results.

In this paper, we propose \textbf{Neural Input Search} (NIS), a novel approach
to find embedding and vocabulary sizes automatically for each discrete feature in the model's input component.
We create a search space consisting of a collection
of Embedding Blocks, where each combination of blocks represents a different vocabulary
and embedding configuration. The optimal configuration is searched for in a single
training run, using a reinforcement-learning algorithm like ENAS~\cite{enas}. Moreover, we propose a novel type of embedding, which we call \textbf{Multi-size
Embedding (ME)}. ME allows allocating larger embedding vectors to
more common or predictive feature items, and smaller vectors to less common or predictive ones.
This is in contrast to a commonly employed approach, which we call Single-size Embedding (SE),
where the same-sized embeddings is used across all items in the vocabulary. We argue that SE is an inefficient use of the model’s capacity and training data. This is because that we need a large embedding dimension for frequent or highly predictive items to encode their nuanced relation with other items,
but training good embeddings of the same size for long tail items may take too many epochs due to their rarity in the training set.
And when training data is limited, large-sized embeddings for rare items can overfit.
With ME, given the same model capacity, we can cover more items in the vocabulary, while reducing the required training data size and
computation cost for training good embeddings for long tail items.

We demonstrate the effectiveness of NIS at finding good configurations of
vocabulary and embedding sizes for both SEs and MEs through experiments on two common types of recommendation problems,
namely retrieval and ranking, using data collected from
our company’s products. In our experiments, NIS is able to automatically find configurations
that result in  $6.8\%$ relative improvement on Recall@1 and $1.8\%$ on ROC-AUC over well established
manually crafted baselines in a single training run.

%% file: related_work.tex

Neural Architecture Search (NAS) has been an active
research area since \cite{zoph2017}, which takes a Reinforcement Learning
approach that requires training thousands of candidate models to
convergence. Due to its resource intensive nature, a lot of
research has focused on developing cheaper NAS
methods. One active research direction is to design a large model that
connects smaller model components, so that different candidate architectures can be
expressed by selecting a subset of the components. The
optimal set of components (and thus the architecture) is learned in a single training run.
For exmaple, ENAS (\cite{enas}) uses a controller to sample the
submodels, and SMASH (\cite{smash}) generates weights for sampled networks using
a hyper-network. DARTS (\cite{liu2018darts}) and SNAS (\cite{snas}) takes a
differentiable approach by representing the connection as a weight, which is
optimized with backpropagation. A similar approach in combination of
ScheduledDropPath (\cite{zoph2018}) on the weights is taken in \cite{bender2018}
and \cite{proxylessnas}. Luo et al. \cite{nao} takes another approach by mapping the neural
architectures into an embedding space, where the optimal embedding is learned
and decoded back to the final architecture.

Another research direction is to reduce the size of the search space.
\cite{real2018, Zhong_2018_CVPR, Liu_2018_ECCV, cai2018} propose
searching convolution cells, which are later stacked
repeatedly into a deep network. Zoph et al. \cite{zoph2018} developed the NASNet architecture and showed the cells learned from
smaller datasets can achieve good results even on larger datasets
in a transfer learning setting. MNAS~\cite{mnas} proposed a search space
comprised of a hierarchy of convolution cell blocks, where cells in different
blocks are searched separately and thus may results in different
structures.

Almost all previous NAS research works have focused on finding the optimal
representation learning component for image/video understanding problems. For
large scale recommendation problems, great results have also been reported by
leveraging advanced representation learning components, such as CNN
(\cite{kim2016}, \cite{oord2013}), RNN (\cite{bansal2016}, \cite{donkers2017}),
etc. However, the input component, although contains a great portion of model
parameters due to large-sized embeddings, has been frequently designed
heuristically across industry, such as YouTube (\cite{youtube_paper}), Google
Play (\cite{wide_deep}), Netflix (\cite{netflix}), etc. Our work, to the best of
our knowledge, for the first time brings automated neural network design into
the input component for large scale recommendation problems.

%% file: neural_input_search.tex

\subsection{Definitions and Notations}
We assume that the model input consists of a set of categorical features $\mathcal{F}$. Each input example can contain any number of values per feature. For each feature $F$, we have a list of its possible values, sorted in decreasing order of frequency of occurrence in the dataset. This list implicity maps each feature value to an integer: we refer to this list as a vocabulary. An embedding variable $E$ is a trainable matrix. If it's shape is $v \times d$, then $v$ is referred to as the vocabulary size and $d$ as the embedding dimension. For any $0 \leq i < v$, we use $E[i]$ to refer to the $i^{th}$ row the embedding matrix $E$, i.e. the embedding vector of the $i^{th}$ item within the vocabulary. Throughout the paper, we use $\mathcal{C}$ to refer to our `memory budget', the total number of floating point values the embedding matrices of the model can use. A $v \times d$ shaped embedding matrix uses $v \times d$ values.

\subsection{Neural Input Search Problems}

We start with introducing our first proposed Neural Input Search problem based on the regular embedding matrix, which we call Single-size Embedding:

\paragraph{Single-size Embedding (SE)}

A single-size embedding is a regular embedding matrix with shape $v \times d$, where each of the $v$ items within the vocabulary is represented as an $d$-dimensional vector. As stated in Section~\ref{intro}, most previous works use SEs to represent discrete features, and the value of $v$ and $d$ for each feature is selected in a heuristic manner, which can be suboptimal. Below we propose a Neural Input Search problem, namely NIS-SE, for automatically finding the optimal SE for each feature, and the approach for solving this problem is introduced later in Section~\ref{optimize}.

\begin{problem}[NIS-SE]
\label{prob:regular-vocab}
Find a vocabulary size $v_F$ and embedding dimension $d_F$ for
each $F \in \mathcal{F}$ to maximize the objective function value of the resulting neural network, subject to:
$$\sum_{F \in \mathcal{F}} v_F \times d_F \leq \mathcal{C}$$
\end{problem}

The problem involves two trade-offs:
\squishlist
\item Memory budget between features: More useful features should get a higher budget.
\item Memory budget between vocabulary size and embedding dimension within each feature.
\squishend

A large vocabulary for a feature gives us higher coverage, letting us include tail items as input signal. A large embedding dimension
improves our predictions for head items, since head items have more training data and larger embeddings can encode more nuanced information. SE makes it difficult to simultaneously obtain high coverage and high quality embeddings within the memory budget. To conquer this difficulty, we introduce a novel type of embedding, namely Multi-size Embedding.

\paragraph{Multi-size Embedding (ME)}

Multi-size Embedding allows \textit{different items in the vocabulary to have different sized embeddings.} It lets us use large embeddings for head items and small embeddings
for tail items. It makes sense to have fewer parameters for tail items as they have lesser training data. The vocabulary and embedding size for a variable is now given by a Multisize Embedding Spec (MES). A MES is a list of pairs: $[(v_1, d_1), (v_2, d_2),\cdots (v_M, d_M)]$ for any $M \geq 1$ such that $v_m \in [1, v] \; \forall \; 1 \leq m \leq M$ and $d_1 \geq d_2 \geq \cdots d_M \geq 1$. This can be interpreted as: the first $v_1$ most frequent items have embedding dimension $d_1$, the next $v_2$ frequent items have embedding dimension $d_2$, etc. The total vocabulary size is $v = \sum_{m=1}^{M}v_m$. When $M=1$, an ME is equivalent to an SE.

Instead of having only one embedding matrix $E$ like in a SE, we create one embedding matrix $E_m$ of shape $v_m \times d_m$ for each $1 \leq m \leq M$. Moreover, a trainable projection matrix $P_m$ of shape $d_m \times d_1$ is created for each $1 \leq m \leq M$, which maps a $d_m$-dimensional embedding to a $d_1$-dimensional space. This facilitates downstream reduction operations to be conducted in the same $d_1$-dimensional space. Define $V_0 = 0$ and $V_m = \sum_{i=1}^{m} v_i$ for $1 \leq m \leq M$ to be the cumulative vocabulary size for the first $m$ embedding matrices, then the ME for $k^{th}$ item in the vocabulary $e_k$ is defined as  

$$e_k = E_{m_k}[k - V_{{m_k}-1}] P_{m_k}$$

where $m_k \in \{1, \cdots, M \} $ is chosen such that $k \in [V_{{m_k} - 1}, V_{m_k})$, and clearly $e_k$ is $d_1$-dimensional. We remind the readers that $E[i]$ represents the $i^{th}$ row of the matrix $E$. 

With an appropriate MES for each feature, ME is able to achieve high coverage on tail items and high quality representation of head items at the same time. However, finding the optimal MSE for all features manually is very hard, necessitating an automated approach for searching the right MESs. Below we introduce the Neural Input Search problem with Multi-size Embedding, namely NIS-ME, and the approach for solving this problem is introduced later in Section~\ref{optimize}.

\begin{problem}[NIS-ME]
\label{prob:multisize-vocab}
Find a MES $[(v_{F_1}, d_{F_1}), (v_{F_2}, d_{F_2}),\cdots (v_{F_{M_F}}, d_{F_{M_F}})]$ for each $F \in \mathcal{F}$ to maximize the objective function value of the resulting neural network, subject to:
$$\sum_{F \in \mathcal{F}} \sum_{i=1}^{M_F} v_{F_i} \times d_{F_i} \leq \mathcal{C}$$
\end{problem}

\begin{figure}
\centering
\begin{subfigure}{.4\textwidth}
\centering
\includegraphics[width=\linewidth]{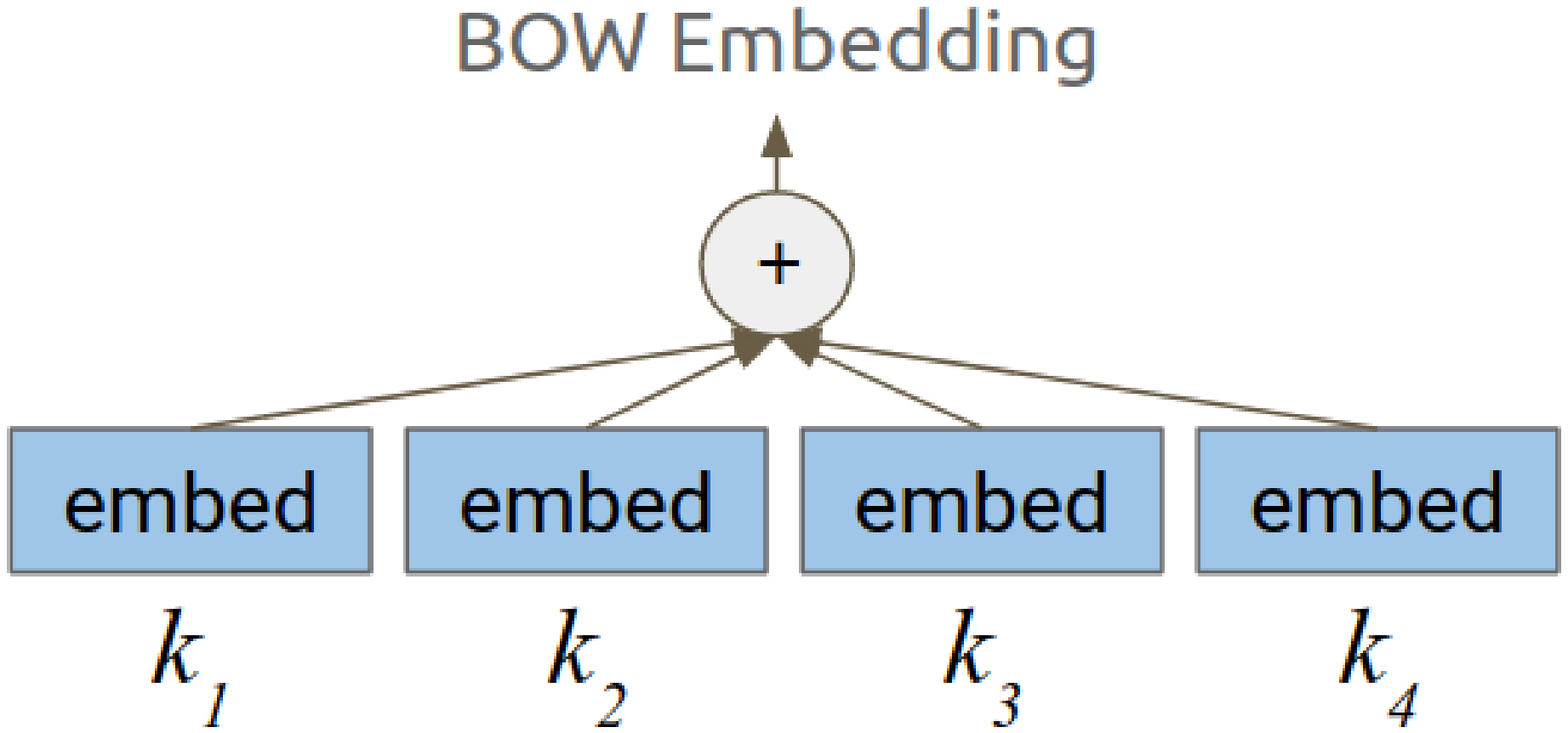}
\caption{}
\label{fig:bow}
\end{subfigure}
\hspace{4pt}
\begin{subfigure}{.4\textwidth}
\centering
 \includegraphics[width=\linewidth]{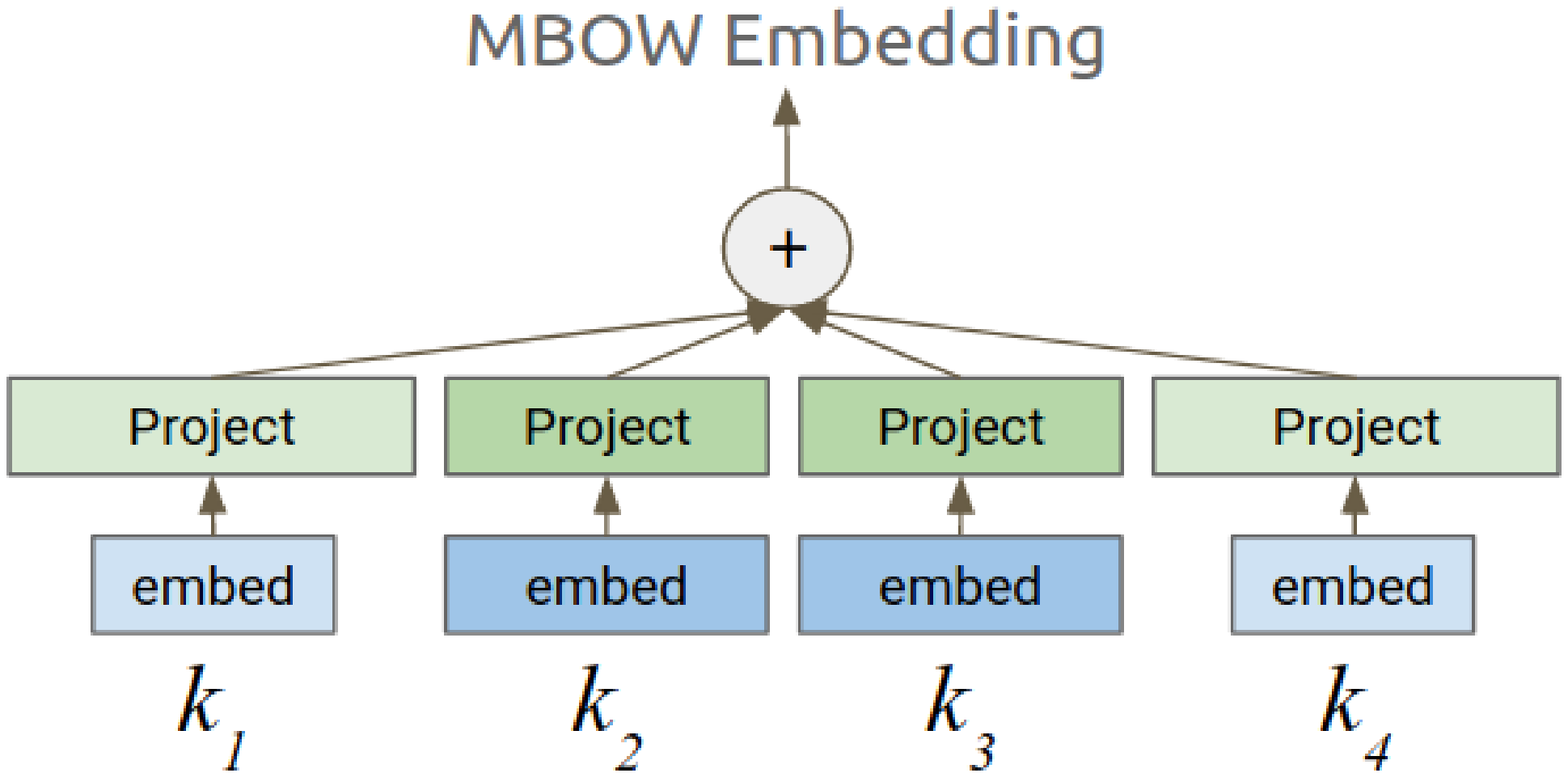}
\caption{}
\label{fig:mbow}
\end{subfigure}
\vspace{-1pt}
\caption{An example of BOW based on SE and ME. (a) BOW with SE: $4$ items from the feature vocabulary are assigned with same sized embeddings, followed by a sum operator. (b) BOW with ME: $k_2$ and $k_3$ are assigned with larger sized embeddings than the other $2$ items. These $4$ embeddings are projected to the same space and summed.}
\vspace{-10pt}
\label{fig:bowvsmbow}
\end{figure}

MEs can be used as a direct replacement for SEs in any model that uses embeddings. Typically, given a set of vocabulary IDs $K$, each element in $K$ is mapped to its corresponding SE, followed by one or more reduce operations to these SEs.  For example, a commonly used reduction operation is bag-of-words (BOW), where the embeddings are summed or averaged. To see how MEs can directly replace SEs in this case, the ME version of BOW, which we call MBOW, is given by:

$$\sum_{k \in K} e_k = \sum_{k \in K}  \left ( E_{m_k}[k - V_{m_k-1}]  P_{m_k}\right )$$

where the MEs are summed. This is illustrated in Figure~\ref{fig:bowvsmbow}. Note that for the $k$'s whose $m_k$ 's are equal, it is more efficient to sum the embeddings before applying the projection matrix.

\subsection{Neural Input Search Approach}
\label{optimize}
We now detail our method for solving Problems~\ref{prob:regular-vocab} and
\ref{prob:multisize-vocab}. As stated in the introduction, most large scale
recommendation models are very expensive to train; it is desirable to solve each of these
problems in one training run. To achieve this goal, we leverage a variant of
ENAS (\cite{enas}): We develop a novel search space in the input component of the model,
 which contains the
SEs or MEs we want to search over. A separate controller is used to make
\textit{choices} to pick an SE or ME for each discrete feature in each step. These selected
SEs or MEs are trained in together with the rest of the main model (excluding the controller). In
addition, we use the feedforward
pass of the main model to compute a reward (a combination of accuracy and memory cost, detailed in
Section~\ref{sec:reward}) of the controller's choices, and the reward is used to train the
controller variables using
the A3C (\cite{a3c}) policy gradient method.

\subsubsection{Search Space}
\label{sec:sparse_voc_emb}

We now describe the search space, which is a key novel ingredient of our work.

\begin{figure}
\centering
\begin{subfigure}{.3\textwidth}
\centering
\includegraphics[width=1.0\linewidth]{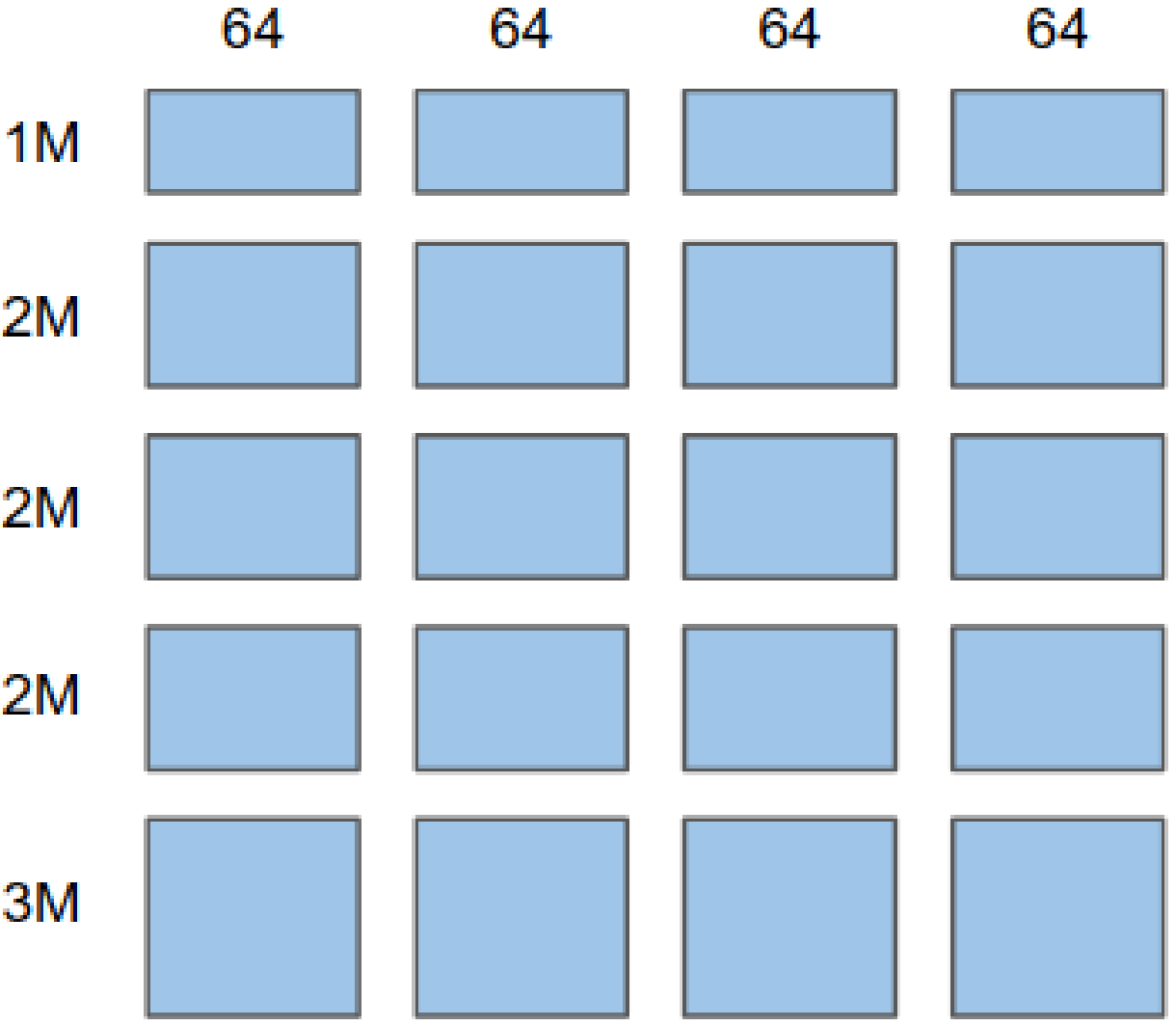}
\caption{}
\label{fig:grid}
\end{subfigure}
\hspace{4pt}
\begin{subfigure}{.3\textwidth}
\centering
\includegraphics[width=1.0\linewidth]{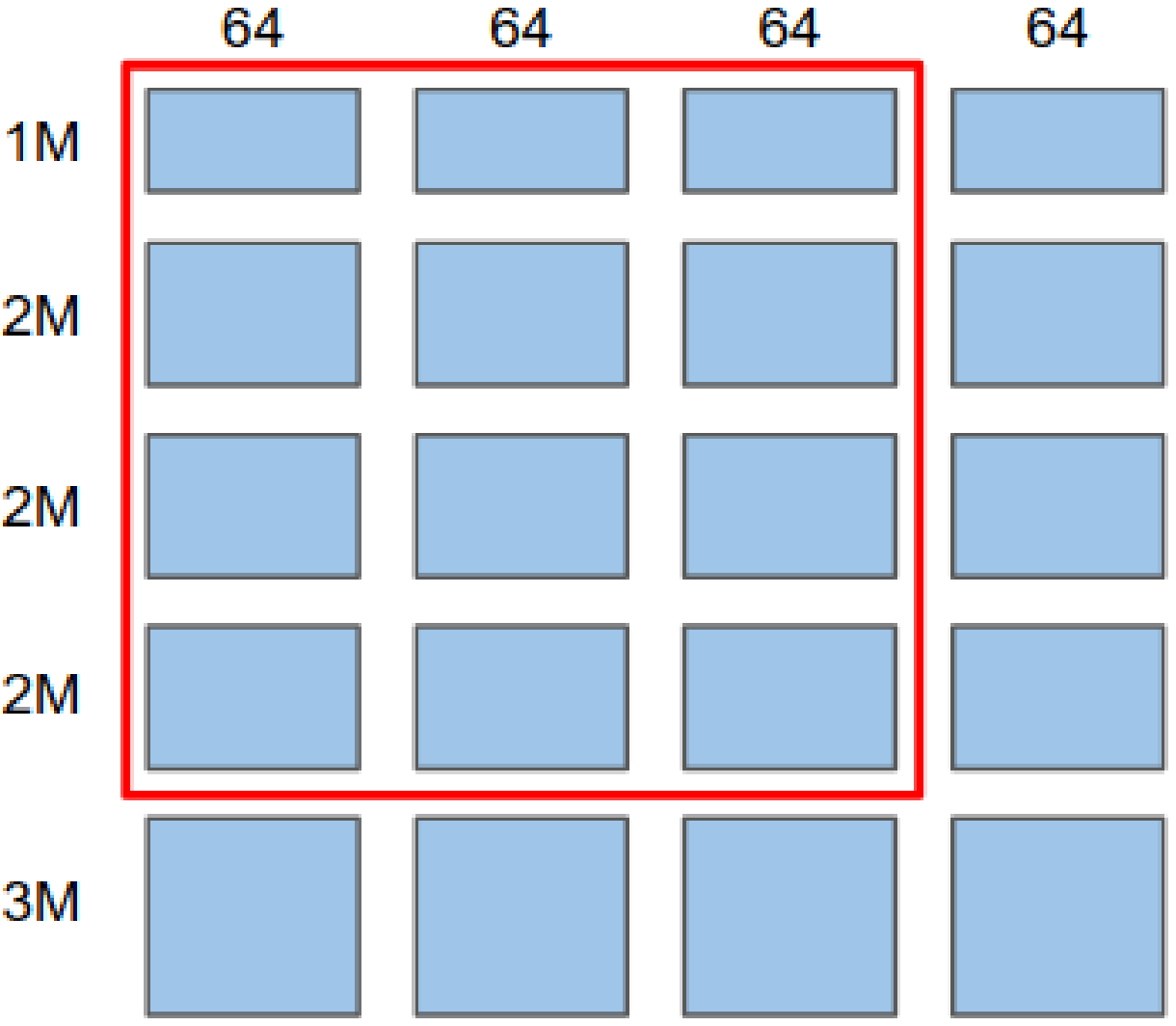}
\caption{}
\label{fig:grid_se}
\end{subfigure}
\hspace{4pt}
\begin{subfigure}{.3\textwidth}
\centering
 \includegraphics[width=1.0\linewidth]{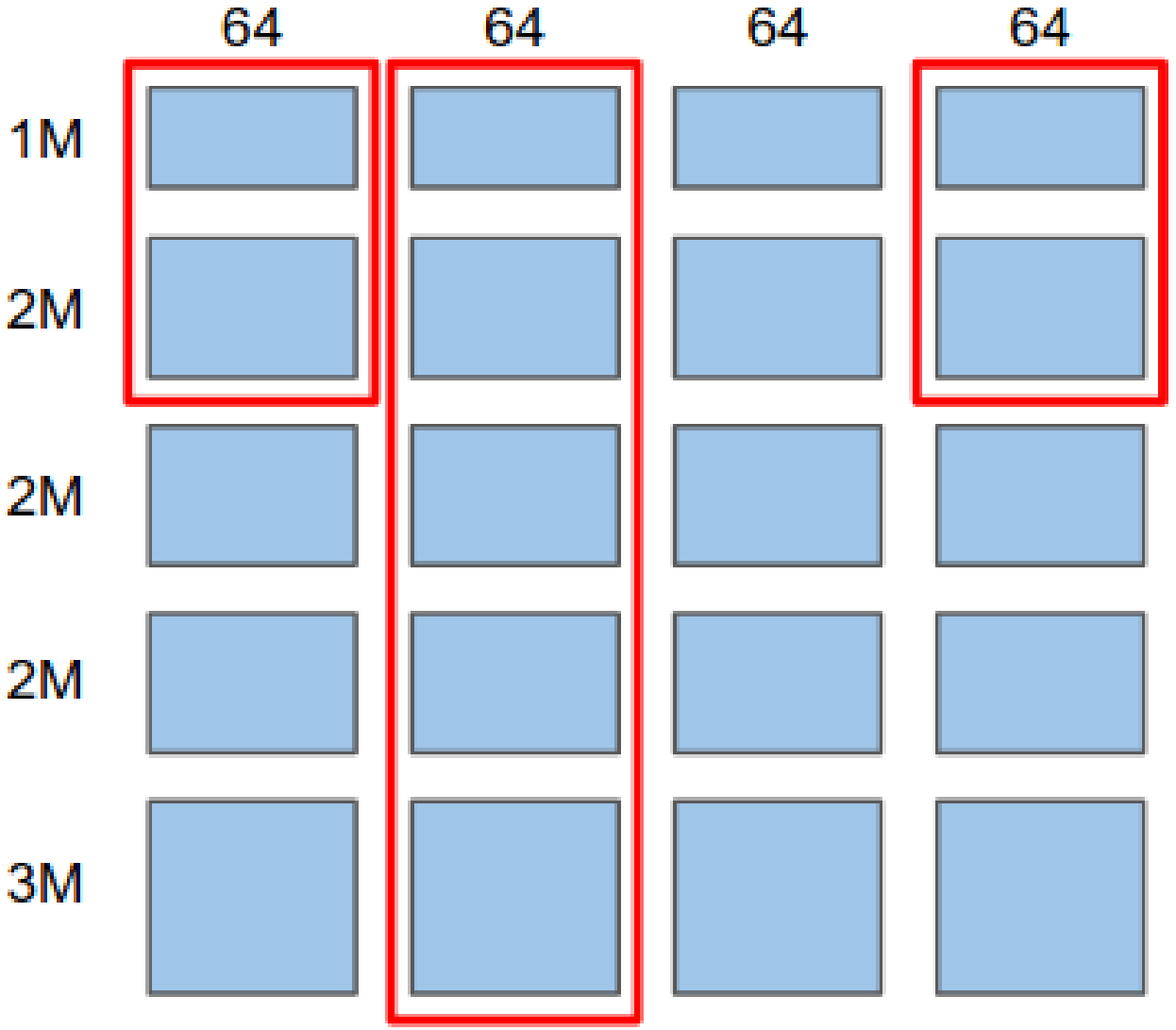}
\caption{}
\label{fig:grid_multisize}
\end{subfigure}
\vspace{-3pt}
\caption{An example of Embedding Blocks and controller choices. (a) An embedding matrix of size $10M \times 256$ is discretized into $20$ Embedding Blocks. (b) The controller samples a $7M \times 192$ sized SE for a training step. (c) The controller samples an ME whose MSE is $[(3M, 192), (7M, 64)]$ for a training step. The first $3M$ items have $192$ dimensional embeddings while the rest $7M$ have $64$ dimensional embeddings.}
\end{figure}

\paragraph{Embedding Blocks}
For a given feature $F \in \mathcal{F}$ with vocabulary size $v$, we create a grid of
$S \times T$ matrices with $S > 1$ and $T > 1$, where the $(s, t)$-th matrix $E_{s, t}$ is of size
$\bar{v}_s \times \bar{d}_t$, such that $v = \sum_{s=1}^S \bar{v}_s$, and $d = \sum_{t=1}^T \bar{d}_t$.
Here $d$ is the maximum allowed embedding size for any item within the
vocabulary. We call these matrices Embedding Blocks.
This can be thought as discretizing an embedding matrix of size $v \times d$
into $S \times T$ sub-matrices. As an example, suppose $v = 10\textrm{M}$
(`$\textrm{M}$' stands for million)
and $d = 256$, we may discretize the rows into five chunks: $[1\textrm{M},
2\textrm{M}, 2\textrm{M}, 2\textrm{M}, 3\textrm{M}]$, and discretize the columns
into four chunks: $[64, 64, 64, 64]$, which results in $20$ Embedding Blocks, as
illustrated in Figure~\ref{fig:grid}. Moreover, a projection matrix $\bar{P}_t$ of size $\bar{d}_t \times d$ is
created for each $t = 1, \cdots, T$, in order to map each $\bar{d}_t$ dimensional embedding to a common
$d$ dimensional space for facilitating downstream reduction operations.  Clearly we should have $\bar{v}_s >> d$ for all $s$. The Embedding Blocks are the
 building blocks of the search space that allow the controller to
sample different SEs or MEs at each training step.

\paragraph{Controller Choices}
The controller is a neural network that samples different SEs or MEs from softmax probabilities. Its exact behavior depends on whether we are optimizing over SEs or MEs. Below we describe the controller's behavior on one feature $F \in \mathcal{F}$, and drop the $F$ subscript for notational convenience. 

\textbf{SE:} To optimize over SEs, at each training step, the controller samples one $(\bar{s},
\bar{t})$ pair from the set $\{(s, t) \mid 1 \leq s \leq S,\: 1 \leq t \leq T \} \; \cup \; \{(0,
0)\}$. For a selected $(\bar{s}, \bar{t})$, only Embedding Blocks $\{ E_{s, t} \mid 1 \leq s \leq
\bar{s}, 1 \leq t \leq \bar{t} \}$ are involved in that particular training step. Therefore, the
controller effectively picks an SE, such as the one within the red rectangle in
Figure~\ref{fig:grid_se}, which represents an SE of size $5M \times 192$. The embedding of the $k^{th}$ item in the vocabulary in this step is calculated as

$$e_k = \sum_{t = 1}^{\bar{t}}E_{s_k, t}[k - \bar{V}_{s_k-1}] \bar{P}_t$$

for all $k < \bar{V}_{\bar{s}}$, where $\bar{V}_0 = 0$, $\bar{V}_s = \sum_{i=1}^s \bar{v}_i$ is the 
cumulative vocabulary size, and $s_k \in \{ 1, \cdots, S \}$ such that $\bar{V}_{s_k-1} \leq k < \bar{V
}_{s_k}$. Define $\bar{D}_t = \sum_{i =1}^t d_t$ to be the cumulative embedding size, it is clear that
 $e_k$ is equivalent to using a $D_{\bar{t}}$-dimensional embedding to represent the
 $k^{th}$ item followed by a projection to a $d$-dimensional space, where the project matrix $P$ is the
concatenation of $\{ \bar{P}_1, \cdots, \bar{P}_{\bar{t}} \} $ along the rows.
Any item whose vocabulary id $k \geq \bar{V}_{\bar{s}}$ is considered as
out-of-vocabulary and is handled specially; a commonly employed approach is
using zero vector as their embedding. The corresponding memory cost (the number
of parameters) induced by this choice of SE is therefore computed as $C =
\bar{V}_{\bar{s}} \times \bar{D}_{\bar{t}}$ (the projection matrix cost is ignored,
since $\bar{v}_s >> d$ for all $s$).

If the pair $(0, 0)$ is selected in a training step, it is equivalent to
removing the feature from the model. Thus the zero embedding is used
for all items of this feature within this training step, and the corresponding
memory cost is $0$. As the controller explores different SEs, it's trained based
on the reward induced by each selection, and eventually converges to the optimal
one, as described in Section~\ref{training}. If it converges to the pair $(0, 0)$, it means this
feature should be removed.

\textbf{ME:} When optimizing over MEs, instead of making a single choice, the
controller makes a sequence of $T$ choices, one for each $t \in {1, \cdots, T}$.
Each choice is an $\bar{s}_t \in \{ 1, \cdots, S \} \cup \{ 0 \}$. If $\bar{s}_t > 0$, 
only Embedding Blocks $\{ E_{s, t} \mid 1 \leq s \leq \bar{s}_t \}$ are involved in that particular training step.
Similarly, if $\bar{s}_t = 0$, it means the whole $\bar{d}_t$-dimensional
embedding is removed for all items within the vocabulary.
Therefore, the controller picks a custom subset (not just a subgrid) of Embedding Blocks, which comprises an MES. This is visually illustrated in Figure~\ref{fig:grid_multisize}, where the first $64$-D
embeddings are utilized by the first $3$M items, the second $64$-D embeddings
are utilized by all of the $10$M items, the third $64$-D embeddings are not used
by any item, while the last $64$-D embeddings have the same utilization as the first $64$-D
embeddings. As a result, the first $3$M items in the vocabulary are allocated with $192$ dimensional embeddings,
while the last $7$M items are assigned with only $64$ dimensional embeddings. In other words, an MES $[(3M, 192), (7M, 64)]$
is realized at this training step.

Mathematically, let $\mathcal{T}_s = \{ t \mid E_{s, t} \; \textrm{is selected}\}$, then the embedding
of the $k^{th}$ item in the vocabulary in this step is calculated as

$$e_k = \sum_{t \in \mathcal{T}_{s_k}} E_{s_k, t}[k - \bar{V}_{s_k-1}] \bar{P}_t$$

for all $k < v$ whose corresponding $\mathcal{T}_{s_k}$ is non-empty, and $e_k$ is an zero vector if $\mathcal{T}_{s_k}$ is empty.
The calculation of memory cost is straightforward: $C = \sum_{t=1}^T \bar{d}_t \times \bar{V}_{\bar{s}_t}$.

\subsubsection{Reward}
\label{sec:reward}
As the main model is trained with the controller's choices of SEs or MEs, the controller is trained with the reward calculated from feedforward passes of the main model on validation set examples. Our reward can be written as $O - C_L$, where $O$ represents the (potentially non-differentiable) objective that we want to optimize, and $C_L$ is cost-loss, a regularization term to force the controller to keep the memory cost within our budget.

\textbf{Objective:} There are two different types of problems that are commonly encountered for recommendation tasks, namely retrieval problems and ranking problems (\cite{youtube_paper}).

Retrieval problems aim at finding the $N$ most relevant items out of a potentially very large vocabulary $v$, given the model's input. $N$ is usually in the hundreds and $v$ is in millions. This is usually achieved by a softmax layer with $v$ neurons, and the $N$ items with the highest softmax probability are used as the results. The objective commonly optimized for is the model's Recall@1. However, since $v$ is large, computing the exact Recall@1 is too expensive to do once per controller training step. We need a cheap proxy of Recall@1. One possibility is to use sampled softmax loss. However, we observed that this is not a good proxy for Recall@1: using very large vocabularies with very small embeddings gives the best sampled softmax loss values, but not the best Recall@1. Instead, we approximate Recall@1 with Sampled Recall@1, i.e. only use the sampled negatives when calculating the recall. Thus Sampled Recall@1 is the fraction of times the logit of the true label was higher than the logits of all the sampled negative labels. We observe that Sampled Recall@1 is a good proxy for Recall@1, and we use it as the $O$ term of our reward for retrieval problems. As Sampled Recall@1 can be calculated for each validation example, given a batch of $b$ examples, the controller can make $b$ different choices, each of which gets trained based on their own reward.

Ranking problems aim at finding the best ranking of a set of items. Such problems involve binary labels (e.g. if the video is watched or not) trained with cross entropy loss. A widely used objective for ranking is the Area Under the Receiver Operating Characteristic Curve (ROC-AUC). However, ROC-AUC can only be computed from a collection of $a$ examples. Therefore, given a batch of $b$ examples, the controller can only make $b / a$ choices, each of which should apply to $a$ examples and result in $b / a$ rewards. The controller will thus explore different choices slower and potentially converge slower in this setting. An alternative is to use the negative cross entropy loss as the objective. Since it can be calculated for each example, the controller can explore different choices with fewer examples. However, we observe  that the controller converges to better results when $O$ is ROC-AUC.

\textbf{Cost Loss:} In Section~\ref{sec:sparse_voc_emb} we defined a cost term $C_F$ based on the choice of the controller (we dropped the subscript $F$ in Section~\ref{sec:sparse_voc_emb} to avoid cluttered notation). We compute the total cost $C = \sum_{f \in \mathcal{F}} C_F$, and define the cost-loss as $C_L = \max(\frac{C}{\mathcal{C}} - 1, 0)$. We remind the reader that $\mathcal{C}$ is the pre-defined memory budget. Note that the cost-loss can be combined with other regularization losses too, e.g. to limit the number of floating point operations used by the model.

\subsubsection{Training}
\label{training}
As stated above, the main model is trained in a regular way using training set examples, where sampled softmax loss is used for retrieval problems and cross entropy loss is used for ranking problems. In addition, we use validation set examples to compute rewards (Section~\ref{sec:reward}), and use the A3C algorithm (\cite{a3c}) to train the controller to maximize the reward.

\textbf{Warm up Phase:} If we start training the controller from step $0$, we get a vicious cycle where the Embedding Blocks not selected by the controller don't get enough training and hence give bad rewards, resulting in them being selected even less in future. To prevent this, the first several training steps consist of a \textit {warm-up phase} where we train all the Embedding Blocks and leave the controller variables fixed. The controller variables are initialized randomly, so the initial controller makes approximately uniformly random choices. The warm up phase ensures that all Embedding Blocks get some training. After the warm up phase, we switch to training the main model and the controller in alternating steps using A3C.

\textbf{Baseline:} As part of the A3C algorithm, we use a baseline network to predict the expected reward prior to each controller choice (but using the choices that have already been made). The baseline network has the same structure as the controller network, but has its own variables, which are trained alongside the controller variables using validation set. Then we subtract the baseline from the reward at each step to compute the advantage, which is used to train the controller.

%% file: experiments.tex

We conduct experiments on two large scale recommendation problems, one for retrieval  and another one for ranking; both are based on real data collected from our company's products.

\paragraph{Query Suggest Retrieval Problem}
This problem is to suggest the next query that the user would like to type in one of our company's Search products, given the last query they issued. The $20$ million most commonly issued queries are used in this experiment; in other words, we want to retrieve a small set of queries that the user would like to type from these $20$ million queries. The input features to the model include full query, query unigrams, bigrams and trigrams from the previous query. We used SE to represent each of the $4$ features, and the SEs are concatenated and fed into the representation learning component of the model, which contains $3$ fully connected hidden layers with ReLU activation function. The output layer is a softmax layer with $20$ million neurons, each of which is associated with a unique query from the label query vocabulary. Our total memory budget is $\mathcal{C} = 2560M$. For the baseline, we tried different combinations of  $v$ and $e$ such that $4 \times v \times e = \mathcal{C}$, using a $v \times e$ embedding for each feature. We used the best performing model as our baseline.

To study the performance of NIS, we used it to find the optimal SE for each of the $4$ features with the same total memory budget $\mathcal{C}$. We also constructed a model with MEs being used as replacement of SEs, while the rest of the model (i.e. the representation learning component and the output layer) is the same. NIS is used to find the optimal MEs for all features. Sampled Recall@1 is used as objective for the controller, where $5000$ negative examples are sampled from the $20$ million vocabulary.

\paragraph{App Install Ranking Problem}
This problem aims at ranking a set of Apps based on the likelihood they will be installed, where the data comes from one  of our company's App store products. This dataset consists (Context, App, Label) tuples, where the Label is either $0$ or $1$, indicating if the App is installed or not. A total of $20$ discrete features are used to represent the Context and App, such as App ID, Developer ID, App Title, etc. The vocabulary size of the discrete features varies from hundreds to millions. Similar to the retrieval problem, SEs of the $20$ features are concatenated and fed into $3$ fully connected layers. Cross entropy loss is used for this ranking problem. For this problem, the App store product already constructed a highly optimized baseline with the corresponding vocabulary size and embedding dimension for each SE. We used the same configuration as our baseline model.

Similar to the retrieval problem, we used NIS to find the optimal SEs given the same memory budget as the baseline model. Moreover, a second model with all SEs being replaced by MEs are constructed, and we again used NIS to find the optimal MEs for all features. Here, the objective for the controller is ROC-AUC, where each controller decision is applied to $100$ validation set examples, and the ROC-AUC is calculated from these $100$ examples.

In all our experiments, $20$ Embedding Blocks are constructed for each feature, with $\bar{v}_s$'s being $[0.1v, 0.2v, 0.2v, 0.2v, 0.3v]$, where $v$ is the total vocabulary size of the feature, and $\bar{d}_t$'s being $[0.25d, 0.25d, 0.25d, 0.25d]$, where $d = 32 \cdot \lceil v^{0.35} / 32 \rceil$, a heuristic value that works well in practice. Here $\lceil \cdot \rceil$ is the ceiling operator. Note that there is nothing prevent setting $d$ to a larger value and discretize it into more buckets, if there is doubt about the effectiveness of the heuristically selected $d$. Each data set is split into $70\%$, $20\%$ and $10\%$ for training, validation and testing.

\begin{table}
  \caption{Comparison between NIS and baseline on the Query Suggest Retrieval Problem}
  \label{table:yt}
  \centering
  \begin{tabular}{cccc}
    \toprule
    Model    & Cost (Million Floats) & Recall@1 (\%)  & Recall@5 (\%) \\
    \midrule
    Baseline & 2560 & 16.0 & 24.2  \\
    NIS-SE & 2560 & 16.5 & 24.3  \\
    NIS-ME & 2560 & 17.1 & 25.7  \\
    \bottomrule
  \end{tabular}
\end{table}

\begin{table}
  \caption{Comparison between NIS and baseline on the App Install Ranking Problem}
  \label{table:play}
  \centering
  \begin{tabular}{ccc}
    \toprule
     Model    & Cost (Million Floats) & AUC  (\%) \\
    \midrule
    Baseline & 12  & 72.2    \\
    NIS-SE & 12  & 73.0   \\
    NIS-ME & 12  & 73.5    \\
    \bottomrule
  \end{tabular}
\end{table}

We report the experimental results for these two problems in Table~\ref{table:yt} and Table~\ref{table:play}. It can be seen that the SEs searched by our NIS approach outperforms the baseline model in both of the two problems, evidenting that NIS is able to automatically find much better vocabulary size and embedding dimension in SE setting, comparing to the approach of choosing these hyper-parameters heuristically. Moreover, both of the baselines involve training one model from scratch for each candidate SE configuaration, which is computationally very expensive.  In comparison, all our optimal SEs are found in only one training run, which is a much more efficient approach.

In addition, the sophistication of MEs make it difficult to configure MEs manually for each feature. Our experimental results show that the MEs automatically searched by our NIS approach even outperformed the optimal SEs. Compare to the baseline, our approach achieves $6.8\%$ and $6.1\%$ relative improvement on Recall@1 and Recall@5 for the retrieval problem, and $1.8\%$ relative improvement on ROC-AUC for the ranking problem. This not only empirically evidented that MEs are more efficient representations of discrete features than SEs within a memory budget, but also demonstrated that NIS is an efficient approach for finding the optimal MEs that result in superior perfomance than manually configured vocabulary sizes and embedding dimensions.

%% file: conclusion.tex

We presented \textbf{Neural Input Search} (NIS), a technique for automatically searching the optimal vocabulary and embedding sizes
in the input component of a model. We also introduced \textbf{Multi-size Embedding} (ME), a novel type of embedding that
achieves high coverage of tail items while keeping accurate representation for head items. We demonstrated the effectiveness of NIS and ME with
experiments on large scale retrieval and ranking problems. Our approach received a relative improvement of $6.8\%$ on Recall@1 and $1.8\%$ on ROC-AUC in
only one training run, without increasing the total number of parameters in the model.